\documentclass{bmvc2k}

\usepackage{multirow}
\usepackage[ruled,vlined]{algorithm2e}
\usepackage{tikz}
\usepackage{amsmath}

\newsavebox{\tempbox}
\DeclareMathOperator*{\argmax}{arg\,max}

\title{Adversarial Concurrent Training: Optimizing Robustness and Accuracy Trade-off of Deep Neural Networks}

\def\etal{\emph{et al}\bmvaOneDot}

\begin{document}

\addauthor{Elahe Arani*}{elahe.arani@navinfo.eu}{1}
\addauthor{Fahad Sarfraz*}{fahad.sarfraz@navinfo.eu}{1}
\addauthor{Bahram Zonooz}{bahram.zonooz@gmail.com}{1}

\addinstitution{
 Advanced Research Lab\\
 NavInfo Europe\\
 Eindhoven, The Netherlands
}

\runninghead{Arani, Sarfraz, Zonooz}{Adversarial Concurrent Training}

\maketitle

\begin{abstract}
Adversarial training has been proven to be an effective technique for improving the adversarial robustness of models.
However, there seems to be an inherent trade-off between optimizing the model for accuracy and robustness.
To this end, we propose {\it Adversarial Concurrent Training} (ACT), which employs adversarial training in a collaborative learning framework whereby we train a robust model in conjunction with a natural model in a minimax game.
ACT encourages the two models to align their feature space by using the task-specific decision boundaries and explore the input space more broadly.
Furthermore, the natural model acts as a regularizer, enforcing priors on features that the robust model should learn.
Our analyses on the behavior of the models show that ACT leads to a robust model with lower model complexity, higher information compression in the learned representations, and high posterior entropy solutions indicative of convergence to a flatter minima.
We demonstrate the effectiveness of the proposed approach across different datasets and network architectures.
On ImageNet, ACT achieves 68.20\% standard accuracy and 44.29\% robustness accuracy under a 100-iteration untargeted attack, improving upon the standard adversarial training method's 65.70\% standard accuracy and 42.36\% robustness.
\end{abstract}

\section{Introduction}
\label{sec:intro}
Deep neural networks (DNNs) have emerged as a predominant framework for learning multiple levels of representation, with higher levels representing more abstracts aspects of the data~\cite{bengio2013deep}. The better representation has led to the state-of-the-art performance in many challenging tasks in computer vision~\cite{krizhevsky2012imagenet, voulodimos2018deep}, natural language processing~\cite{collobert2008unified, young2018recent} and many other domains~\cite{pierson2017deep, heaton2017deep}. However, despite their pervasiveness, recent studies have exposed the lack of robustness of DNNs to various forms of perturbations~\cite{szegedy2013intriguing, hendrycks2019benchmarking, gu2019using}. In particular, adversarial examples which are imperceptible perturbations of the input data carefully crafted by adversaries to cause erroneous predictions pose a real security threat to DNNs deployed in critical applications~\cite{kurakin2016adversarial}. 

The intriguing phenomenon of adversarial examples has garnered a lot of attention in the research community~\cite{yuan2019adversarial} and progress has been made in both creating stronger attacks to test the model's robustness~\cite{goodfellow2014explaining, moosavi2015deepfool, carlini2017towards, xiao2018spatially} as well as defenses to these attacks~\cite{madry2017towards, Zhang2019theoretically, lamb2019interpolated}. However, \citet{athalye2018obfuscated} show that most of the proposed defense methods rely on obfuscated gradients which is a special case of gradient masking and lowers the quality of the gradient signal causing the gradient-based attack to fail and give a false sense of robustness. They observe adversarial training~\cite{madry2017towards} as the only effective defense method. The original formulation of adversarial training, however, does not incorporate the clean examples into its feature space and decision boundary. On the other hand, \citet{jacobsen2018excessive} provide an alternative viewpoint and argue that the adversarial vulnerability is a consequence of narrow learning, resulting in classifiers that rely only on a few highly predictive features in their decisions.
We have not yet developed a full understanding of the major factors that contribute to adversarial vulnerability in DNNs and consequently, the optimal method for training robust models remains an open question.

A recent variant of adversarial training, TRADES~\cite{Zhang2019theoretically}, adds a regularization term on top of the standard cross-entropy loss which forces the model to match its embeddings for the clean example and the corresponding adversarial example. However, there might be an inherent tension between the objective of adversarial robustness and that of standard generalization~\cite{tsipras2018robustness}. Therefore, combining these optimization tasks into a single model and forcing the model to completely match the feature distributions of the adversarial and clean examples may lead to sub-optimal solutions. We, therefore, hypothesize that considering the optimization for adversarial robustness and generalization as two distinct yet complementary tasks and encouraging more exhaustive exploration of the input and parameter space can lead to better solutions.
 
\begin{figure}[tb]
\centering
\includegraphics[clip, trim=0cm .3cm 0cm .3cm, width=13cm]{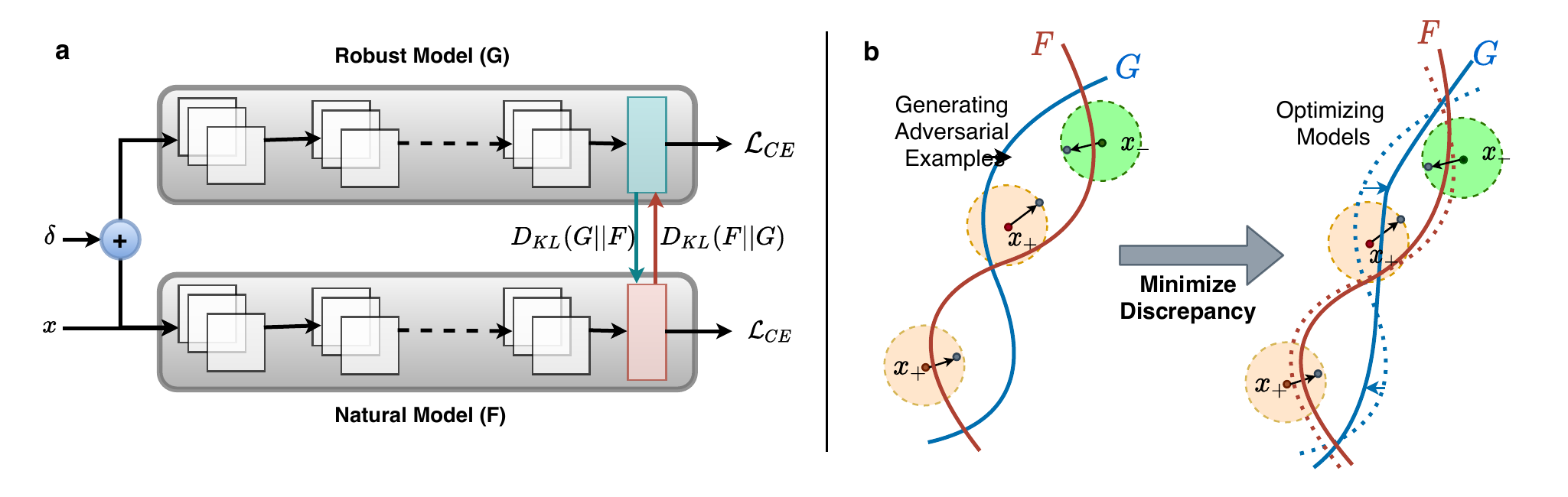}
\caption{a) Schematic of our proposed method.
b) An overview of ACT on a binary classification problem. $x_{+}$ and $x_{-}$ indicate data samples for positive and negative classes respectively whereas circle indicates the allowed $\epsilon$-bound. First, adversarial examples are generated by identifying discrepancy regions between $G$ and $F$. The arrow in the circles shows the direction of the adversarial perturbation and the circles show the perturbation bound. In a subsequent step, the discrepancy between the models is minimized. This effectively aligns the decision boundaries and pushes them further from the examples. Best viewed in color.}
\label{fig:act_schematic}
\end{figure}

In this paper, we propose adversarial concurrent training (ACT) for training a robust model in conjunction with a natural model in a collaborative manner (Fig.\ref{fig:act_schematic}a). The goal is to utilize the task-specific decision boundaries to align the feature space of the robust and natural model in order to learn a more extensive set of features that are less susceptible to adversarial perturbations. To this end, ACT closely intertwines the training of a robust and natural model by involving them in a minimax game inside a closed learning loop. The adversarial examples are generated by determining regions in the input space where the discrepancy between the two models is maximum. In the subsequent step, each model minimizes a supervised learning loss which optimizes the model on its specific task in addition to a mimicry loss that aligns the two models. Our formulation consists of bi-directional knowledge distillation between the clean and adversarial domain, enabling them to collectively explore the input and parameter space more extensively. Furthermore, the supervision from the natural model acts as a regularizer which effectively adds a prior on the learned representations and leads to semantically meaningful features that are less susceptible to off-manifold perturbations introduced by adversarial attacks.

We empirically test the efficacy of our proposed approach and show that ACT provides a better trade-off between robustness and generalization across different datasets (CIFAR-10, CIFAR-100~\cite{krizhevsky2010cifar} and ImageNet \cite{krizhevsky2012imagenet}) and network architectures (ResNet~\cite{he2015deep} and WideResNet~\cite{zagoruyko2016wide}). 
Our further analyses show that ACT learns a lower complexity model with higher posterior entropy solutions, indicative of convergence to flatter minima. While standard adversarial training reduces the information compression in the learned representations compared to standard training~\cite{lamb2019interpolated}, our method shows higher information compression than even standard training. The empirical results coupled with desirable characteristics of models trained with ACT demonstrates the effectiveness of concurrent training for adversarial robustness.
Our results also demonstrate the versatility of ACT to different datasets and network architectures which makes the method applicable across a variety of application

\section{Related Work}
The discovery of adversarial examples \cite{szegedy2013intriguing} has garnered a lot of interest from the research community.
Researchers have proposed various forms of defense methods which include detecting the adversarial examples \cite{grosse2017statistical, feinman2017detecting}, applying non-linear pre-processing and transformations on the input image, using ensemble method \cite{strauss2017ensemble, tramer2017ensemble, buckman2018thermometer,bhagoji2017dimensionality}, regularization techniques \cite{zantedeschi2017efficient, jakubovitz2018improving, summers2019improved} and training on adversarial examples \cite{goodfellow2014explaining, madry2017towards, Zhang2019theoretically,lamb2019interpolated}.
However, Athalye \etal \cite{athalye2018obfuscated} showed that most of the proposed defense methods rely on gradient obfuscation, lowering the quality of the gradient signal, to give a false sense of robustness.
They found adversarial training to be an effective method after addressing the issue of gradient obfuscation. 
Nevertheless, the increase in robustness comes at the cost of generalization.
A number of studies even argue that there is an inherent trade-off between robustness and generalization and consider them as contradictory goals \cite{tsipras2018robustness, ilyas2019adversarial, su2018robustness, Zhang2019theoretically}.
Ilyas \etal \cite{ilyas2019adversarial} consider the adversarial vulnerability to be a direct consequence of the model's sensitivity to well generalizing features which are highly predictive yet brittle.
Jacobsen \etal \cite{jacobsen2018excessive} provide an alternative perspective on adversarial vulnerability and show that DNNs are also excessively invariant to task relevant changes in the input image.
They attribute this to narrow learning resulting from the insufficiency of the standard cross-entropy loss to incentivize explaining all class dependent aspects of the input.

On other end, collaborative learning which provides additional supervision signals, has been effective in increasing the robustness to different noise types.
Knowledge distillation \cite{hinton2015distilling} has been shown to be a general-purpose training paradigm which is more robust to common challenges in the real-world datasets \cite{sarfraz2020knowledge}. Han \etal~\cite{han2018co} use two networks to filter different types of errors introduced by noisy labels.
Hendrycks \etal~\cite{hendrycks2019using} show that self-supervision can improve the robustness of the model to adversarial examples, label corruption, and common input corruptions.
Based on the aforementioned findings, we hypothesize that adversarial training within a collaborative framework that encourages the model to explore the input and parameter space more extensively can be instrumental in further improving the robustness gains of the standard adversarial training method.

\section{Adversarial Concurrent Training}
In this section, we first present the overall idea and intuition behind the proposed method and how it aims to address some of the shortcomings of standard adversarial training and then formally define the method and introduce the loss functions used for training the model.

\subsection{Proposed Method}\label{act_overview}
Standard adversarial training~\cite{madry2017towards} involves generating an adversarial example $x'$ for each clean example $x$ and then subsequently training the model to assign the same label $y$ to $x'$ using cross-entropy loss. Adversarial training has been proven to confer robustness to the model. However, the standard formulation has a few shortcomings. The model does not receive any pair information indicating that $x'$ is the adversarial counterpart of $x$ and therefore fails to utilize the semantic similarity between the adversarial and clean examples for learning an optimal embedding. The objective function does not involve explicitly minimizing the generalization on clean examples which can lead to overfitting to the adversarial domain. Furthermore, the model is not incentivized to incorporate all class-dependent features of the input into its decision boundary which leads to narrow learning. 
One approach is to combine the generalization and adversarial robustness loss into one objective function \cite{Zhang2019theoretically,kannan2018adversarial}.
However, the goal of adversarial robustness is different from standard generalization~\cite{tsipras2018robustness}. Therefore, combining these two optimization tasks together into a single model and completely matching the feature distributions of the adversarial and clean examples could cause tension between the two tasks and leads to sub-optimal solutions.

We hypothesize that treating the optimization for adversarial robustness and generalization as distinct yet complementary tasks in a way that encourages more exhaustive exploration of the input and parameter space can lead to better solutions. To this end, we propose Adversarial Concurrent Training (ACT) which entails training an adversarially robust model in conjunction with a natural model in a collaborative manner (Fig.\ref{fig:act_schematic}a). The goal is to utilize the task specific decision boundaries to align the feature space of the robust and natural model in order to learn a more extensive set of features which are less susceptible to adversarial perturbations. ACT closely intertwines the training of a robust and natural model by involving them in a minimax game inside a closed learning loop. The adversarial examples are generated by identifying regions in the input space where the discrepancy between the robust and natural model is maximum. In the subsequent step, the discrepancy between the two models is minimized in addition to optimizing them on their respective tasks.

\begin{algorithm}[tb]
\caption{Adversarial Concurrent Training Algorithm}
\label{algo:act}
\textbf{Input:} Dataset $D$, Balancing factor $\alpha$, Learning rate $\eta$, Batch size $m$\\
\textbf{Initialize:} $G$ and $F$ parameterized by $\theta$ and $\phi$\\
\While{Not Converged}{
1: Sample mini-batch: ${(x_1, y_1), ... , (x_m,y_m)} \sim D$\\
2: Compute adversarial examples:\\
~~~~$\delta^*=\argmax_{\delta\in S}\mathcal{L}_{G}(\theta,\phi,\delta)$\\
3: Compute $\mathcal{L}_G(\theta,\phi,\delta^*)$ (Equation \ref{eq:loss_G})\\
~~~~Compute $\mathcal{L}_F(\theta,\phi,\delta^*)$ (Equation \ref{eq:loss_F})\\
4: Compute stochastic gradients and update the parameters:\\
~~~~$\theta^*\gets\theta-\eta\frac{\partial\mathcal{L}_{G}}{\partial\theta}$\\
~~~~$\phi^*\gets\phi-\eta\frac{\partial\mathcal{L}_{F}}{\partial\phi}$
}
\Return{$\theta$* and $\phi$*}
\end{algorithm}

Our approach has a number of advantages. The adversarial perturbations, generated by identifying regions in the input space where the two models disagree, can be effectively used to align the two models. This alignment coupled with pushing the two decision boundaries away from the data samples leads to smoother decision boundaries (Fig.\ref{fig:act_schematic}b).
Updating the models based on the disagreement regions combined with optimization on distinct tasks ensures that the two models do not converge to a consensus, and the method does not reduce to self-training.
Furthermore, the supervision from the natural model acts as a noise-free reference for regularizing the robust model. This effectively adds a prior on the learned representations which encourages the model to learn semantically relevant features in the input space.
This combined with the requirement on the robust model's prediction to be stable within the epsilon bound encourages the model to select semantically relevant features with stable behavior over a larger region.

\subsection{Formulation}\label{formulation}
We formulate our proposed method, ACT, as a concurrent training of an adversarially robust model $G$ parametrized by $\theta$ and a natural model $F$ parametrized by $\phi$ (see Fig.\ref{fig:act_schematic}a). Each model is trained with two losses: a task specific loss and a mimicry loss.
The standard cross-entropy loss ($\mathcal{L}_{CE}$) is used as the task specific loss and the Kullback-Leibler Divergence (${D}_{KL}$) is used as the mimicry loss to align the output distributions of the models.
The robust model minimizes the convex combination of the cross-entropy loss on adversarial examples and the ${D}_{KL}$ between the output distributions of the robust model on adversarial examples and the natural model on clean examples.
\begin{equation}\label{eq:loss_G}
\begin{split}
\mathcal{L}_G(\theta,\phi,\delta)= &(1-\alpha)\mathcal{L}_{CE}(G(x + \delta;\theta),y)+\alpha D_{KL}(F(x;\phi)||G(x + \delta;\theta))
\end{split}
\end{equation}
where the perturbation $\delta$ is sampled from a set of allowed perturbations $S$ bounded by $\epsilon$. The tuning parameter $\alpha \in[0,1]$ plays key role on balancing the importance of task specific and alignment errors. The natural model uses a similar loss function which minimizes the cross-entropy loss on clean examples.
\begin{equation}\label{eq:loss_F}
\begin{split}
\mathcal{L}_F(\theta,\phi,\delta)=&(1-\alpha)\mathcal{L}_{CE}(F(x;\phi),y)+\alpha D_{KL}(G(x+\delta;\theta)||F(x;\phi))
\end{split}
\end{equation}

The training procedures involves first finding the adversarial examples by maximizing the robust model loss $\mathcal{L}_{G}$ with respect to $\delta$ within the set of allowed perturbation $S$, and then subsequently minimizing the loss functions for each model $\mathcal{L}_{G}$ and $\mathcal{L}_{F}$ (see Algorithm \ref{algo:act}). This results in an approximate minimax optimization:
\begin{equation}\label{eq:optim}
\begin{cases}
    \min_{\theta}\mathop{E}_{(x,y)\in D}\max_{\delta\in S}\mathcal{L}_{G}(\theta,\phi,\delta)\\
    \min_{\phi}\mathop{E}_{(x,y)\in D}\mathcal{L}_{F}(\theta,\phi,\delta)
\end{cases}
\end{equation}

Note that the natural model $F$ is only used during training, and at inference, only the robust model $G$ is used. Therefore, ACT does not incur any additional inference cost compared to standard adversarial training.

\section{Empirical Validation}
In this section, we empirically evaluate the the effectiveness of our proposed method and study the characteristics of the models.

\subsection{Experimental Setup}\label{exp_setup}
We evaluate the performance ACT on different datasets (CIFAR-10, CIFAR-100 \cite{krizhevsky2010cifar} and ImageNet \cite{krizhevsky2012imagenet}) and network architectures (ResNet \cite{he2015deep} and WideResNet \cite{zagoruyko2016wide}). For all our experiments, we normalize the images between 0 and 1 and apply random cropping with reflective padding of 4 pixels and random horizontal flip data augmentations. For training, we use stochastic gradient descent (SGD) with 0.9 momentum, 200 epochs, batch size 128, and an initial learning rate of 0.1, decayed by a factor of 0.2 at epochs 60, 120 and 150. Unless explicitly mentioned, the results for ACT refers to the performance of the robust model G. For Madry and TRADES, we follow the training scheme used in \cite{Zhang2019theoretically}. For generating adversarial examples during training, we use the projected gradient decent (PGD) as a universal first order adversary \cite{madry2017towards} with $\epsilon=0.031$, step size $\eta=0.007$, and the perturbation steps $K=10$. For evaluation, we set $\eta = 0.003$ and test for different perturbation steps. For a fair comparison, we use $1/\lambda = 5$ for TRADES which achieves the highest robustness for ResNet-18 in \cite{Zhang2019theoretically}. In our experiments, TRADES achieves both better robustness and generalization than reported in the original work \cite{Zhang2019theoretically}.

We train each method with 3 different random seeds and report the average and one standard deviation performance. $A_{nat}$ refers to standard accuracy on clean examples whereas $A_{rob}$ refers to accuracy on adversarial examples (reported in percentage). 
Unless otherwise stated, $A_{rob}$ shows the worst performance on a PGD-20 attack with 5 random initialization.

\begin{table}[tb]
\center
\resizebox{\columnwidth}{!}{%
\begin{tabular}{|l|l|cccccc|}
\hline
\multicolumn{2}{|l|}{} & 0.1 & 0.3 & 0.5 & 0.7 & 0.9 & 1.0 \\ \hline
\multirow{2}{*}{\begin{tabular}[c]{@{}l@{}}Standard \\ model (F)\end{tabular}} & $A_{nat}$ & 95.29$\pm$0.10 & 95.31$\pm$0.10 & 94.88$\pm$0.13 & 94.42$\pm$0.10 & 90.12$\pm$0.63 & 10.00\\
 & $A_{rob}$ & 3.57$\pm$1.06 & 3.81$\pm$1.29 & 2.90$\pm$0.26 & 3.36$\pm$0.45 & 10.85$\pm$2.04 &  0.00\\ \hline
\multirow{2}{*}{\begin{tabular}[c]{@{}l@{}}Robust\\ model (G)\end{tabular}} & $A_{nat}$ & 85.94$\pm$0.11 & 86.13$\pm$0.14 & 86.33$\pm$0.22 & 86.24$\pm$0.18 & 84.87$\pm$0.91 & 10.00\\
 & $A_{rob}$ & 48.93$\pm$0.28 & 49.40$\pm$0.59 & 50.62$\pm$0.86 & 51.37$\pm$0.41 & 55.12$\pm$1.07 & 0.00\\ \hline
\end{tabular}}
\caption{Effect of $\alpha$ hyperparameter on ACT (ResNet-18 trained on CIFAR-10).
\label{tab:alpha_eff}}
\end{table}

\subsection{Effect of $\alpha$ hyperparameter}\label{alpha_param}
Table \ref{tab:alpha_eff} shows the effect of the balancing factor $\alpha$ on the robustness and generalization of the natural and robust models. 
The extra supervision signal from each of the model affects both the robustness and generalization performance of the models. The adversarial robustness of the robust model generally increases as we increase $\alpha$ value. Interestingly, considerable robustness is transferred to the natural model as well without explicitly being trained on adversarial examples and this transfer increases for higher $\alpha$ values.
For our subsequent experiments, we use $\alpha=0.9$.

\begin{table}[tb]
\center
\resizebox{\columnwidth}{!}{%
    \begin{tabular}{|l|l|l|c|c|c|c|c|}
        \hline
        \multirow{2}{*}{} & \multirow{2}{*}{Dataset} & \multirow{2}{*}{Defense} & \multirow{2}{*}{$A_{nat}$} & \multicolumn{3}{c|}{$A_{rob}$} & Minimum \\ \cline{5-7}
         & & & & PGD-20 & PGD-100 & PGD-1000 & Perturbation\\ \hline \hline
         \multirow{6}{*}{\rotatebox[origin=c]{90}{ResNet-18}} & \multirow{3}{*}{CIFAR-10} & Madry & \bf 85.11$\pm$0.19 & 50.53$\pm$0.01 & 47.67$\pm$0.18 & 47.51$\pm$0.16 & 0.03782$\pm$0.00024\\ 
         & & TRADES & 83.49$\pm$0.38 & 53.79$\pm$0.29 & 52.15$\pm$0.26 & 52.12$\pm$0.26 & 0.04279$\pm$0.00066 \\ 
         & & ACT & 84.33$\pm$0.27 & \bf 55.83$\pm$0.18 & \bf 53.73$\pm$0.19 & \bf 53.62$\pm$0.19 & \bf 0.04454$\pm$0.00069\\ \cline{2-8}
         & \multirow{3}{*}{CIFAR-100} & Madry & 58.36$\pm$0.10 & 24.48$\pm$0.16 & 23.10$\pm$0.20 & 23.02$\pm$0.23 & 0.01961$\pm$0.00010\\ 
         & & TRADES & 56.91$\pm$0.46 & 28.88$\pm$0.16 & 27.98$\pm$0.17 & 27.96$\pm$0.19 & 0.02353$\pm$0.00014\\ 
         & & ACT & \bf 61.56$\pm$0.46 & \bf 31.14$\pm$0.16 & \bf 29.74$\pm$0.15 & \bf 29.71$\pm$0.14 & \bf 0.02462$\pm$0.00017\\ \hline
         \multirow{6}{*}{\rotatebox[origin=c]{90}{WRN-28-10}} & \multirow{3}{*}{CIFAR-10} & Madry & 87.26$\pm$0.20 & 49.76$\pm$0.06 & 46.91$\pm$0.10 & 46.77$\pm$0.06 & 0.04412$\pm$0.00083\\ 
         & & TRADES & 86.36$\pm$0.26 & 53.52$\pm$0.17 & \bf 50.73$\pm$0.18 & \bf 50.63$\pm$0.17 & 0.04714$\pm$0.00018\\ 
         & & ACT & \bf 87.58$\pm$0.16 & \bf 54.94$\pm$0.14 & 50.66$\pm$0.11 & 50.44$\pm$0.13 & \bf 0.05601$\pm$0.00031\\ \cline{2-8}
         & \multirow{3}{*}{CIFAR-100} & Madry & \bf 60.77$\pm$0.16 & 24.92$\pm$0.23 & 23.56$\pm$0.26 & 23.46$\pm$0.24 & 0.02084$\pm$0.00011\\ 
         & & TRADES & 58.10$\pm$0.17 & 28.49$\pm$0.08 & \bf 27.50$\pm$0.23 & \bf 27.44$\pm$0.23 & 0.02395$\pm$0.00011\\ 
         & & ACT & 60.72$\pm$0.18 & \bf 28.74$\pm$0.14 & 27.32$\pm$0.00 & 27.26$\pm$0.01 & \bf 0.02595$\pm$0.00016 \\ \hline
    \end{tabular}}
\caption{Comparison of ACT with prior defense models under various white-box attacks.
\label{tab:comparison1}}
\end{table}

\subsection{Comparison with prior work}\label{comparison}

As our method adapts standard adversarial training in a collaborative learning framework, original formulation by Madry \cite{madry2017towards} is included as baseline. Furthermore, TRADES \cite{Zhang2019theoretically} is included to show the effectiveness of optimization for robustness and generalization as two distinct yet complementary tasks instead of combining them into a single model.
Table \ref{tab:comparison1} shows the effectiveness of ACT across different datasets and network architectures under various white-box attacks. Specifically, for ResNet-18, ACT significantly improves the robustness.
In instances where Madry has better generalization, the difference in the robustness is considerably larger.

\begin{figure}[tb]
    \centering
    \begin{tabular}{cc}
        \includegraphics[width=.47\columnwidth]{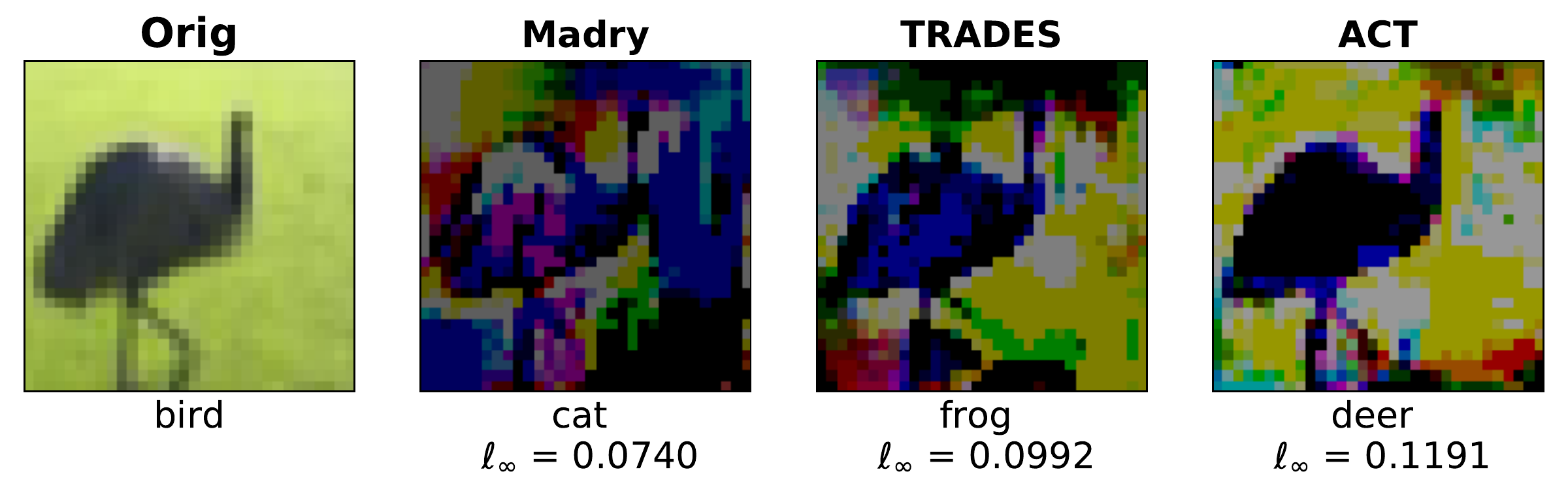}&
        \includegraphics[width=.47\columnwidth]{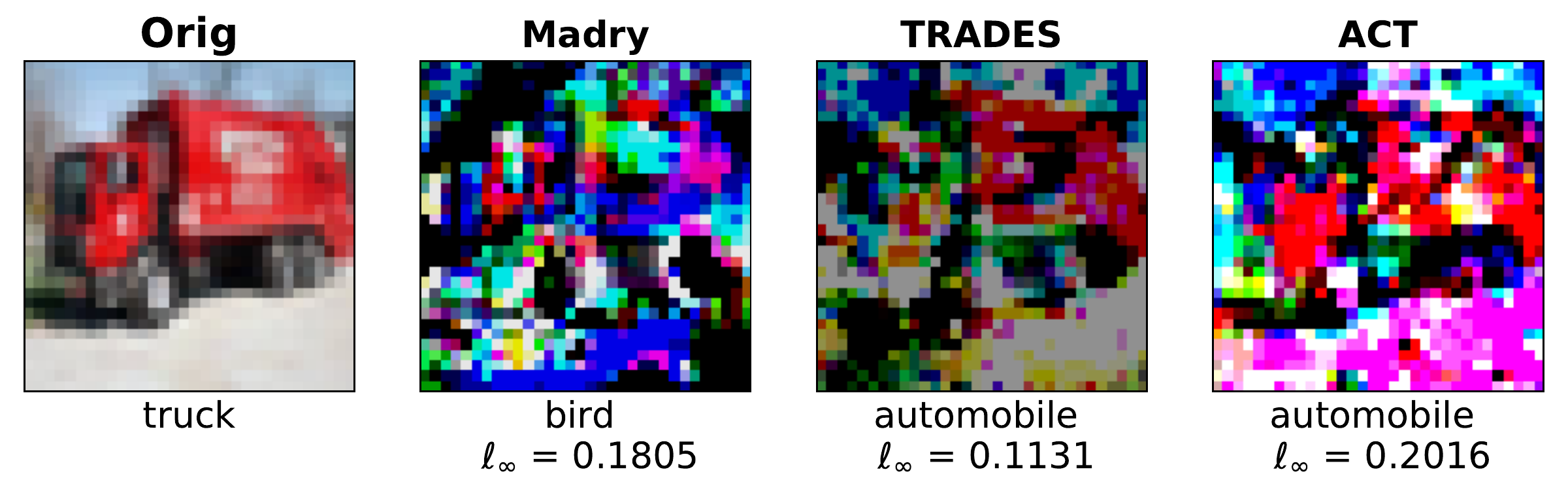}\\
        \includegraphics[width=.47\columnwidth]{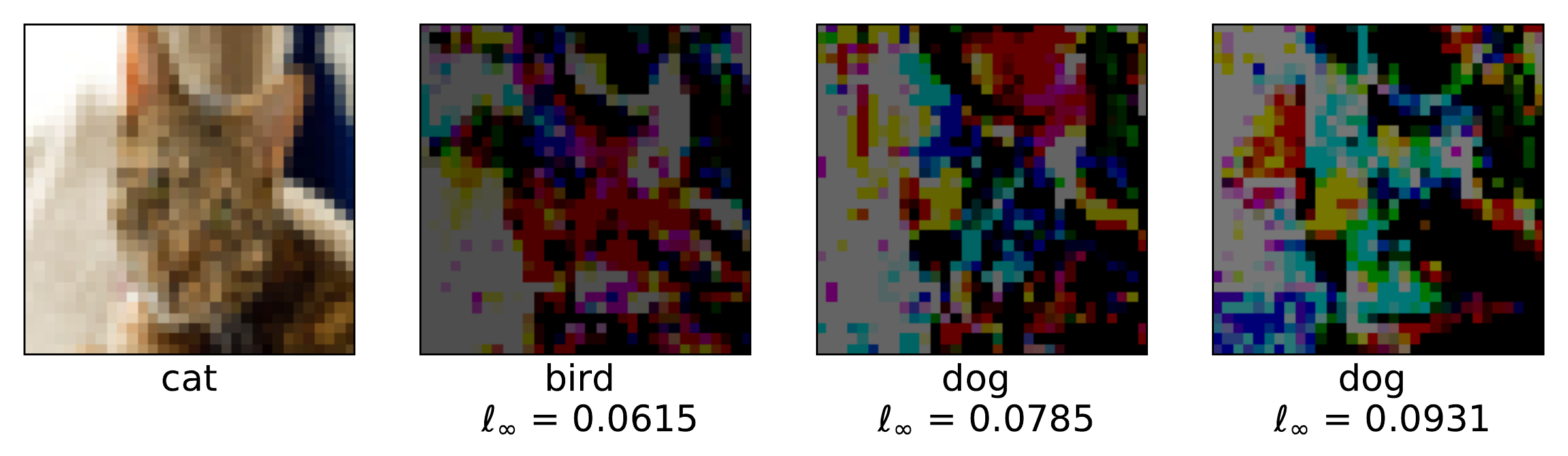}&
        \includegraphics[width=.47\columnwidth]{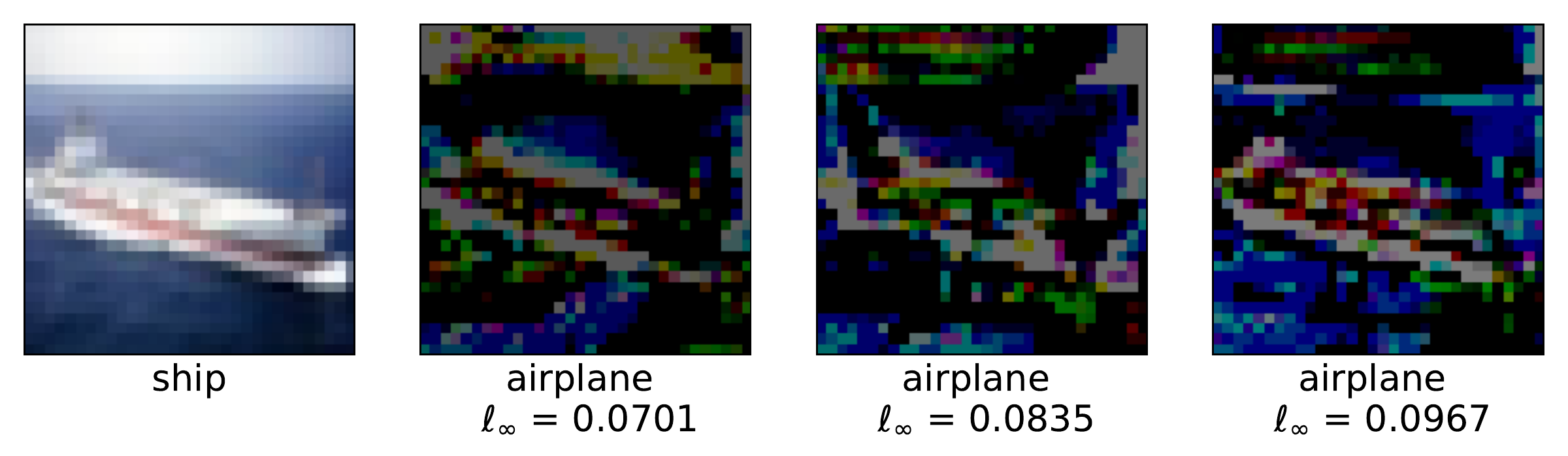}\\
    \end{tabular}
    \caption{Minimum perturbations required to fool the robust models trained with different defense methods on ResNet-18 and CIFAR-10.
    The label of each image shows the predicted class along with the $\ell_\infty$ distance of the adversarial example from the clean example. Note that the perturbations are multiplied by 5 to highlight the visual differences.}
    \label{fig:adversarial_example}
\end{figure}

We also evaluate the average minimum perturbation required to successfully fool the defense methods. We apply the $FGSM^k$ attack in foolbox \cite{rauber2017foolbox} which returns the smallest adversarial perturbation under the $\ell_\infty$ distance. Table \ref{tab:comparison1} shows that ACT consistently requires higher perturbation in images on average across the different datasets and network architectures. Fig.\ref{fig:adversarial_example} provides examples of the required perturbations to fool each defense. ACT requires a higher degree of perturbation in the semantically relevant regions of the image.

We further verify the effectiveness of our method under black-box attacks. 
Table \ref{tab:blackbox} shows the transferability of adversarial examples generated using a PGD-20 attack on the surrogate model to the target models trained with different defense methods. 
ACT shows higher robustness to black-box attacks transferred from Madry and TRADES. 

\savebox{\tempbox}{\begin{tabular}{@{}r@{}l@{\space}}
&\rotatebox[origin=b]{0}{\bf Sur} \\ \rotatebox[origin=t]{0}{\bf Def}
\end{tabular}}
\begin{table}[tb]
    \center
    \resizebox{\linewidth}{!}{%
    \begin{tabular}{|l|cccc|cccc|}
        \hline
        \multirow{2}{*}{\tikz[overlay]{\draw (0pt,\ht\tempbox) -- (\wd\tempbox,-\dp\tempbox);}%
        \usebox{\tempbox}\hspace{\dimexpr 1pt-\tabcolsep}} & \multicolumn{4}{c|}{CIFAR-10} & \multicolumn{4}{c|}{CIFAR-100} \\\cline{2-9}
         & Natural & Madry & TRADES & ACT & Natural & Madry & TRADES & ACT \\\hline\hline
        Madry  & \bf 15.93 & 49.50 & 35.88 & \bf 35.34 & 43.13 & 75.71 & 61.50 & 60.29 \\
        TRADES & 18.23 & 35.84 & 46.49 & 37.11 & 44.25 & 60.80 & 71.20 & \bf 59.97 \\
        ACT    & 16.93 & \bf 33.65 & \bf 35.14 & 44.04 & \bf 40.80 & \bf 57.17 & \bf 57.32 & 68.61 \\\hline
    \end{tabular}}
\caption{Comparison of ACT with prior defenses (Def) under black-box PGD-20 attack. Surrogate models (Sur) are source models that provide gradients for adversarial attacks. Values indicate the success rate of adversarial attack hence a lower number shows higher robustness.}
\label{tab:blackbox}
\end{table}

\subsection{Results on ImageNet}\label{imagenet}
We further demonstrate the effectiveness of our approach on the challenging ImageNet classification task \cite{krizhevsky2012imagenet}.
As our approach is designed for untargeted adversarial training, following prior work on untargeted attacks on ImageNet dataset \cite{shafahi2019adversarial,wong2020fast}
we train a Resnet-50 model to be robust to untargeted PGD attack with the following parameters: $\epsilon=2/255$, $\eta = 1.0$, and $K = 10$. For our experiments, we use four Tesla V100 GPUs with $\alpha=0.5$, batch size of 128 on each GPU, and train for 100 epochs with an initial learning rate of 0.1 decayed by a factor of 0.1 at 30, 60 and 90 epochs. For a fair comparison, we also train Madry \cite{madry2017towards} under the same experimental setup. Table \ref{tab:imagenet} shows that ACT improves both the generalization and robustness over the standard adversarial training method and its faster variants. 
As PGD attack is sensitive to the initial randomization, there can be small fluctuations in the final robustness. Therefore, the results for the different PGD attack essentially show that robustness is maintained even as we increase the number of steps.

\begin{table}[tb]
\center
\begin{tabular}{|l|l|ccc|}
\hline
\multirow{2}{*}{Method} & \multirow{2}{*}{$A_{nat}$} &  \multicolumn{3}{c|}{$A_{rob}$} \\ \cline{3-5}
                                            &                & PGD-10     & PGD-50 & PGD-100 \\ \hline \hline
Madry                                       & 65.70          & 42.13      & 42.29  & 42.36        \\
Free (m=4) \cite{shafahi2019adversarial}    & 64.45          & 43.52      & 43.39  & 43.40   \\
FGSM \cite{wong2020fast}                    & 60.90          & 43.46      & -      & -       \\ 
ACT                                         & \bf 68.20      & \bf 44.06  & \bf 44.24       & \bf 44.29        \\ \hline  
\end{tabular}
\caption{Comparison of ACT with prior defenses under untargeted PGD attack with $\epsilon=2/255$ on the ImageNet dataset.}
\label{tab:imagenet}
\end{table}

\subsection{Gradient obfuscation}\label{grad_obf}
\citet{athalye2018obfuscated} showed that most of the proposed defense methods give a false sense of security by reducing the quality of the gradient signal and that these defenses can be circumvented by using gradient approximation techniques. Therefore, it is important to perform a number of sanity checks to ensure that a proposed adversarial defense does not rely on gradient obfuscation. These checks include ensuring that white-box attacks are at least as strong as black-box attacks and that an unconstrained iterative gradient-based attack with an unlimited number of iterations should be completely successful \cite{lamb2019interpolated}. Our evaluations show that black-box attacks are substantially weaker than the corresponding white-box attacks (Tables \ref{tab:comparison1} and \ref{tab:blackbox}). Table \ref{tab:epsilon_eff} shows that the robustness of the model monotonically decreases as we increase the allowed perturbation level for a PGD-100 attack. This shows that the gradients of our method do not impair the ability of the gradient-based attacks through gradient obfuscation.

\begin{table}[tb]
\centering
\begin{tabular}{|l|cccccccc|}
        \hline
        Defense & 1 & 5 & 10 & 15 & 20 & 25 & 50 & 100 \\ \hline \hline
        Madry  & \bf 81.90 & 63.53 & 36.69 & 16.61 & 7.01 & 3.31 & 0.22 & 0 \\ \hline
        TRADES & 80.23 & 65.27 & 42.51 & \bf 23.35 & 11.17 & 5.40 &	0.29 & 0 \\ \hline
        ACT    & 81.47 & \bf 67.15 & \bf 42.98 & 22.45 & \bf 11.18 & \bf 5.74 & \bf 0.42 & 0 \\ \hline
    \end{tabular}
\caption{Accuracy of ResNet-18 with different $\epsilon$ values and a fix number of steps (PGD-100) conducted on CIFAR-10.
\label{tab:epsilon_eff}}
\end{table}

\subsection {Model complexity}\label{model_complexity}
\begin{figure}[tb]
    \centering
    \includegraphics[width=.42\columnwidth]{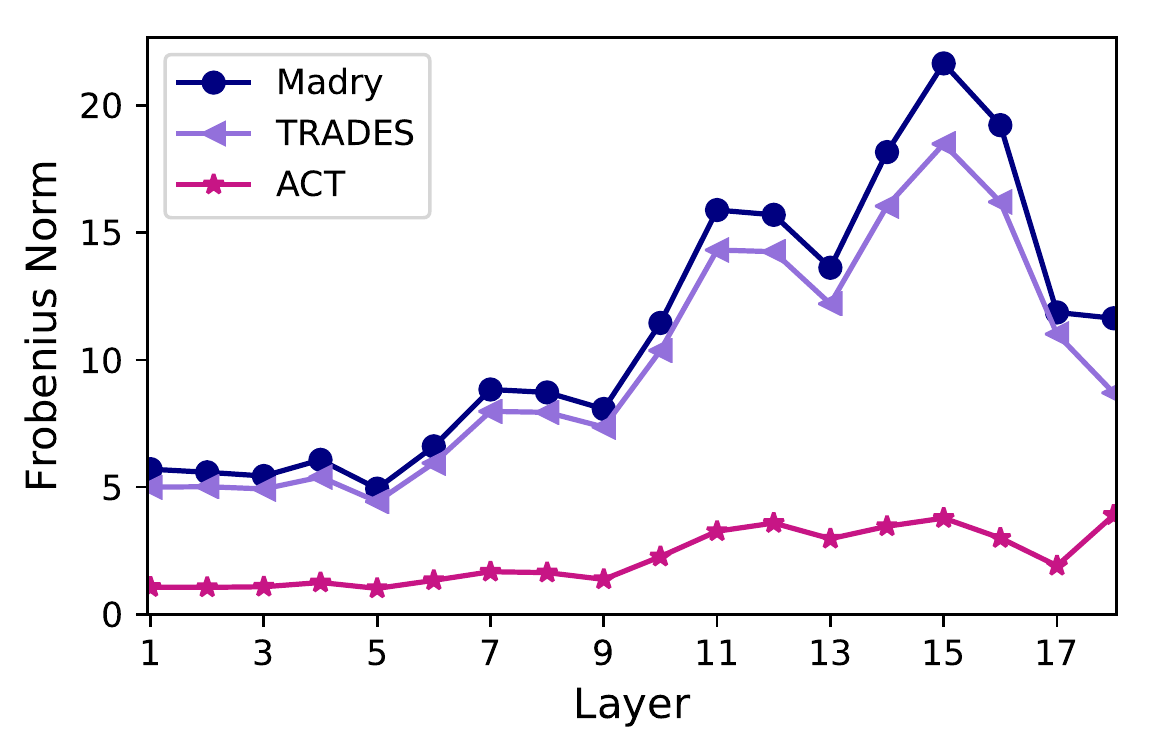} 
    \includegraphics[width=.45\columnwidth]{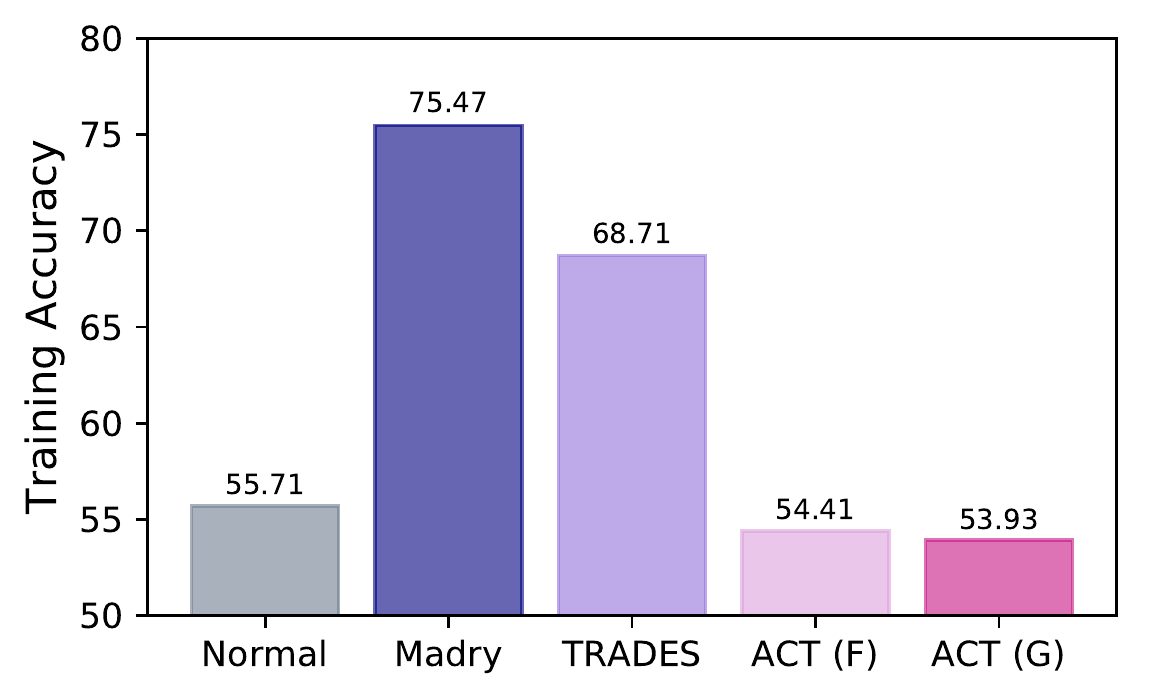}
    \caption{(Left) Comparison of the Frobenius norms of the weights matrix in learnable layers of ResNet-18. (Right) Comparison of the degree to which the different models with frozen learned representation can fit random binary labels.}
    \label{fig:defense_weights_norm}
\end{figure}

The magnitudes of the weights of neural networks can provide an estimate of the model's complexity. Following the analysis presented in \cite{lamb2019interpolated}, we analyze the Frobenius norms of all the weight layers in ResNet-18 for different defense methods trained on CIFAR-10. Fig.\ref{fig:defense_weights_norm} (left) shows that the Frobenius norm of ACT is considerably lower across all the layers. This provides preliminary evidence that ACT trains lower complexity models than standard adversarial training. 

\subsection{Information compression}\label{info_comp}
A number of studies on understanding DNNs from an information theory perspective have shown a relationship between the information compression in the learned features and generalization \cite{tishby2015deep, shwartz2017opening}. They relate stronger compression in DNN's hidden states to a stronger bound on generalization. To study the effect of our proposed method on the compression of information in the learned representation, we follow the analysis performed by \citet{lamb2019interpolated} whereby we freeze the learned representation of the model and study how successful these frozen representations are in predicting fixed random labels. In particular, we add a 2-layer multi-layer perceptron (MLP) network with 400 and 200 neurons on top of the frozen representations of ResNet-18 models trained on CIFAR-10 with different defense methods and fit them on random binary labels. If the model compresses the information well in the learned representations, it will be more difficult to fit the random binary labels. Thereby, lower accuracy shows better information compression. \citet{lamb2019interpolated} showed that standard adversarial training causes the learned representation to be less compressed. To the contrary, Fig.\ref{fig:defense_weights_norm} (right) suggests that both the natural and robust model trained with ACT has more information compression.
Interestingly, the models trained with ACT shows higher information compression compared to standard training (normal). This indicates the efficacy of our proposed approach in capturing more information in the hidden states of the models.

\subsection{Entropy regularization}\label{entropr_reg}
There can be multiple solutions that can fit the training data distribution, but some of these generalize better because of being in wide valleys rather than narrow crevices \cite{chaudhari2019entropy, keskar2016large} whereby the predictions do not change drastically with small perturbations. A number of studies have shown that the tendency towards finding these robust minima can be increased by biasing the DNNs towards solutions with higher posterior entropy \cite{chaudhari2019entropy, pereyra2017regularizing}.

\begin{figure}[tb]
    \centering
    \includegraphics[width=.6\columnwidth]{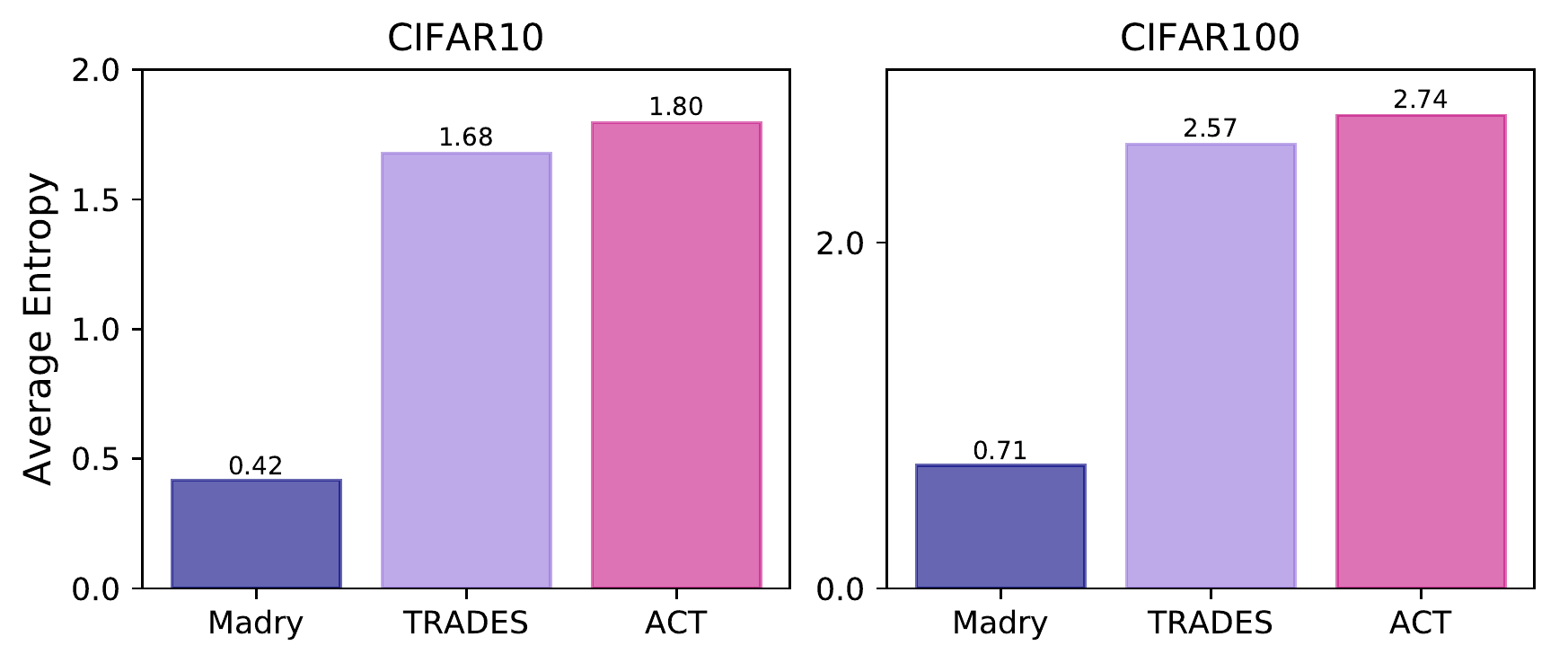}
    \caption{Average entropy over all training samples obtained by a ResNet-18 on CIFAR-10 and CIFAR-100 datasets. ACT converges to higher posterior entropy solutions.}
    \label{fig:mean_posterior_prob}
\end{figure}

We argue that the extra mimicry loss which encourages the model to match the posterior probabilities in ACT has a regularization effect on the logits. The effect is that the model distributes its mass over the secondary classes more uniformly. This can be quantified with the average posterior entropy over the training samples. Fig.\ref{fig:mean_posterior_prob} shows that training with ACT leads to higher posterior entropy solutions. Therefore, the collaborative learning in ACT has a connection to entropy regularisation-based approaches \cite{chaudhari2019entropy, pereyra2017regularizing} to finding wider minima through mutual probability matching on secondary classes.

\section{Conclusion}
We proposed {\it Adversarial Concurrent Training (ACT)} as an efficient approach to training a robust model in conjunction with a natural model. The additional supervision from the natural model allows the robust model to learn richer internal representation which is robust to adversarial perturbations. Our empirical results showed that ACT provides a better trade-off between robustness and generalization across different datasets and network architectures. Furthermore, our analysis suggests that ACT leads to a robust model with lower model complexity, higher information compression in the learned representation, and high posterior entropy solutions indicative of convergence to a flatter minima. The versatility of our proposed approach coupled with the desirable characteristics makes it applicable across a variety of tasks and applications.





\bibliography{references.bib}
\end{document}